# Putting AI Ethics into Practice: The Hourglass Model of Organizational AI Governance

Version February 2023

Matti Mäntymäki, Matti Minkkinen, Teemu Birkstedt, and Mika Viljanen

University of Turku, Finland

**Abstract**

The organizational use of artificial intelligence (AI) has rapidly spread across various sectors. Alongside the awareness of the benefits brought by AI, there is a growing consensus on the necessity of tackling the risks and potential harms, such as bias and discrimination, brought about by advanced AI technologies. A multitude of AI ethics principles have been proposed to tackle these risks, but the outlines of organizational processes and practices for ensuring socially responsible AI development are in a nascent state. To address the paucity of comprehensive governance models, we present an AI governance framework, the hourglass model of organizational AI governance, which targets organizations that develop and use AI systems. The framework is designed to help organizations deploying AI systems translate ethical AI principles into practice and align their AI systems and processes with the forthcoming European AI Act. The hourglass framework includes governance requirements at the environmental, organizational, and AI system levels. At the AI system level, we connect governance requirements to AI system life cycles to ensure governance throughout the system's life span. The governance model highlights the systemic nature of AI governance and opens new research avenues into its practical implementation, the mechanisms that connect different AI governance layers, and the dynamics between the AI governance actors. The model also offers a starting point for organizational decision-makers to consider the governance components needed to ensure social acceptability, mitigate risks, and realize the potential of AI.

# Putting AI Ethics into Practice: The Hourglass Model of Organizational AI Governance

## 1. Introduction

Artificial intelligence (AI) is undergoing a global upswing, and in particular, machine learning solutions based on Big Data have proliferated over the past decade (Dignum, 2019, Chapter 2). The quickly increasing availability of data and technological advances in Big Data processing have led to a resurgence of AI research and use after the "AI winter" of the 1980s and 1990s (Rahwan, 2018). The use of AI has rapidly spread in organizations across sectors such as healthcare (Trocin et al., 2021), policing (Rezende, 2020), and finance (Hua et al., 2019). Private and public sector actors are adopting AI systems to gain process efficiency, enhance their decision speed and quality, and ultimately advance their organizational goals more effectively (Sun & Medaglia, 2019; Taeihagh, 2021; Wirtz et al., 2019). Alongside the awareness of the benefits brought about by AI, there is a growing consensus among researchers and practitioners on the need to tackle the risks and potential harms brought about by advanced AI technologies (Dignum, 2020). For example, potential biases and discrimination in algorithmic recruitment (Fumagalli et al., 2022) have led to calls for the rigorous auditing of AI recruitment systems (Kazim et al., 2021). Even when biases cannot be detected, the opacity and inscrutability of AI systems raise ethical questions (Barredo Arrieta et al., 2020). Biases and transparency invite questions of accountability with regard to discerning networks and relationships of responsibility (Martin, 2019). As AI delves deeper into the everyday lives of citizens, including sensitive areas such as healthcare and personal finance, the systemic risks of AI technologies grow significantly.

The continuing advancement of AI in high-risk application areas, such as healthcare, traffic, and finance, and stakeholders' alertness to its potential risks make the effective governance of AI systems a necessity in the coming years. The growing awareness of AI risks has thus far yielded numerous guidelines on AI ethics principles (Jobin et al., 2019) and increasing regulatory pressure, including a proposal for an AI Act in the European Union (EU) (European Commission, 2021). Aiming to operationalize AI ethics principles, scholars and practitioners have started to discuss organizational and societal AI governance (Dafoe, 2018; Eitel-Porter, 2021; Mäntymäki et al., 2022; Schneider et al., 2020). Only recently, research has started to converge toward explicit definitions of AI governance (Mäntymäki et al., 2022). A summary of the current state of the literature reveals that AI governance comprises tools, rules, processes, procedures, and values that aim to ensure the legally compliant and ethically aligned development and use of AI (Butcher & Beridze, 2019; Gahnberg, 2021; Mäntymäki et al., 2022; Winfield & Jirotka, 2018). While the importance of AI governance has been repeatedly noted (Butcher & Beridze, 2019; Cath, 2018; Gasser & Almeida, 2017; Schmitt, 2021), comprehensive, practice-oriented frameworks for governing AI are few (Benjamins et al., 2019; Eitel-Porter, 2021; cf. Minkkinen, Laine, et al., 2022). Collections, reviews, and syntheses of AI ethics principles are in plentiful supply (Hagendorff, 2020; Jobin et al., 2019), but the outlines of organizational processes and practices that are necessary for ensuring sustainable AI development are in a nascent state.

Typically, AI governance models touch on particular aspects, such as fairness or transparency (Benjamins et al., 2019), and focus on specific stages of system development, such as system design. However, organizations need to govern AI systems over their entire life cycles and consider the complete set of requirements vis-à-vis ethics, legislation, and stakeholders (Laato et al., 2021; Laato, Mäntymäki, Islam, et al., 2022). Moreover, as most organizations cannot tackle complex AI governance problems alone (Minkkinen, Zimmer, et al., 2022), organizations need to understand the different elements of AI governance and their own part in the multi-actor responsible AI ecosystems (Minkkinen, Zimmer, et al., 2022).

To address this paucity of comprehensive governance models, we present an AI governance framework, the hourglass model of organizational AI governance, which targets organizations that develop and use AI systems. The framework is designed to help organizations deploying AI systems to translate ethical AI principles into practice and align their operations with the forthcoming European AI Act.

Our work draws on a wide range of literature. The identification of the focal areas of AI governance was informed by the literature on AI ethics (Hagendorff, 2020; Jobin et al., 2019; Shneiderman, 2020), IT governance (Brown & Grant, 2005; Gregory et al., 2018), IT system life cycles (OECD, 2019), organizational theory (Harjoto & Jo, 2011), and critical algorithm studies (Kitchin, 2017; Ziewitz, 2016). These literature sources provide two kinds of inputs. First, the literature identifies the key features of AI and other algorithmic technologies that pose AI-specific governance problems. These include data-related biases, intractability, and lack of explainability. Second, the literature highlights the risk dynamics present in AI technologies. For its operational components, our framework draws on insights from the IT system life cycle literature to map likely AI development process timelines and phases. Finally, the framework is informed by organizational theory and regulatory and legal studies.

The framework has been developed in collaboration with companies that use AI systems and provide AI consulting. In developing our framework, we paid particular attention to the emerging EU AI legislation. At the time of writing, the EU approach culminated in the proposed EU AI Act (European Commission, 2021) and the ongoing negotiations regarding its final form. In practice, we cross-checked each element of the AI governance framework presented in this paper with the relevant sections of the AI Act proposal available at the time of writing.

We focus on the EU AI Act for two reasons. First, AI policy in the EU has been under active development since 2018 (European Commission, 2018) and is more developed than the legislation in the United States and the state-led system in China. Second, even though we focus on the EU law, other regions may adopt compatible AI regulations in the near future. Thus, organizations that adhere to the EU law will most likely be well-positioned to align themselves with legislation in other regions. In addition to legislation, we focus on soft law and soft ethics that exceed the minimum legal compliance (cf. Floridi, 2018). However, the governance processes discussed in the following sections presuppose that AI systems are used within legal bounds.

The remainder of the paper is structured as follows. The next section introduces the starting points of our AI governance framework and, in particular, its layered approach. Then, we present the framework, visualized as an hourglass, with explanations of each component. The fourth section discusses the theoretical and practical implications of the proposed hourglass model of organizational AI governance.

## 2. Layers of AI Governance

The literature considers AI governance to be a layered phenomenon consisting of distinct levels (Brendel et al., 2021; Gasser & Almeida, 2017; Shneiderman, 2020). Gasser and Almeida (2017) recognize the social and legal layer (norms, regulation, and legislation), the ethical layer (criteria and principles), and the technical layer (data governance, algorithm accountability, and standards). Shneiderman (2020), in turn, identifies three levels of governance: team (engineering practices within teams), organization (safety culture), and industry (independent oversight and trustworthy certification). Brendel et al. (2021) use the common distinction between strategic, tactical, and operational management to discuss levels of organizational decision-making on AI.

The AI governance literature suggests different criteria for the distinct layers. Governance layers can be interpreted as qualitatively different requirements with different logics (Gasser & Almeida, 2017), levels of action and leverage over algorithmic systems (Shneiderman, 2020), and managerial decision-making horizons (Brendel et al., 2021). While all of these distinctions are valid, we propose an organizational AI governance framework that centers on AI systems as IT artifacts that are employed in organizations that are embedded within their operating environments. Each organization may use several AI systems, and each operating environment may host numerous organizations.

This conceptualization of AI governance layers yields a structure whereby the AI system is the concrete governed entity. We conceptualize the AI system as an information technology (IT) artifact that includes AI technologies and is surrounded by a socio-technical system that consists of people, organizations, work systems, and institutions (Dignum, 2019, 2020; Hevner et al., 2004). The organizational and environmental layers constitute the scaffoldings for AI development and use. Fig. 1 illustrates this situation using three distinct layers: environmental, organizational, and AI system. The layers are conceived as qualitatively different, depicting different levels of abstraction.

In addition to this three-layer structure, we consider the bottom AI system layer to be a nested structure due to its complexity. In other words, the AI system layer comprises several governance components and, ultimately, specific processes and tools. Hence, the bottom layer, the AI system, is most directly relevant to the practical implementation of AI governance. Before describing this operational governance layer, we present the hourglass model of organizational AI governance as an overarching framework.

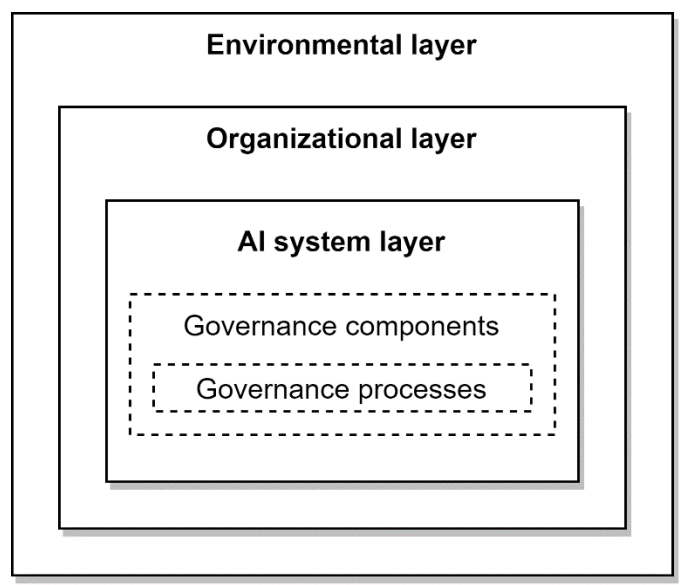

*Fig. 1. The layered structure of AI governance*

## 3. The Hourglass Model of Organizational AI Governance

We propose an *hourglass model of organizational AI governance* to illustrate the AI governance components (Fig. 2). We developed the model with industry partners in a research project and tested it with a large Finnish financial sector company. In the hourglass model, AI governance layers (see Fig. 1 above) are stacked with the environmental level at the top. The organizational level is in the middle, while the operational governance of AI systems is at the bottom.

The hourglass metaphor denotes the flow of governance requirements from the environmental layer to AI systems through the mediating organizational layer. The metaphor also highlights the dynamic nature of AI governance as a continuous activity that translates the normative regulatory, self-regulatory, and stakeholder inputs into operational practices (Seppälä et al., 2021). These translation activities may take place on varying organizational levels and functions, from management to chief AI officers, corporate social responsibility officers, and development teams.

The metaphor of an hourglass thus likens the translation process to the flow of grains of sand (environmental inputs) through the middle of an hourglass (the translation process). AI ethics principles and legal and societal requirements do not automatically ensure responsible use of AI but must be translated by organizations into practicable AI governance processes and mechanisms.

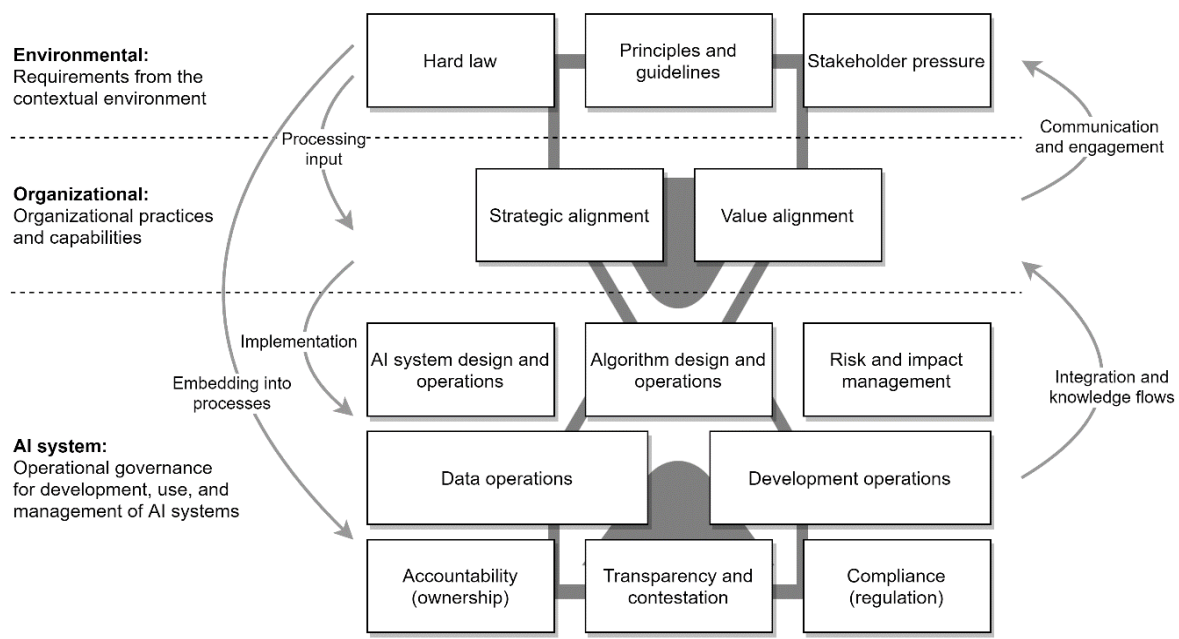

*Fig. 2. The hourglass model of organizational AI governance*

The environmental layer in Fig. 2 refers to inputs from an organization's contextual environment—that is, from societal actors beyond its direct influence (Emery & Trist, 1965). The organizational layer comprises organizational practices and capabilities, and the AI system layer holds the operational governance practices for AI systems. The AI governance layers influence one another both from the top down and the bottom up.

In the top-down direction, environmental inputs are processed by organizations *(processing input)* and implemented in AI systems *(implementation)*. In addition, many legal and ethical requirements directly concern AI systems (e.g., documentation requirements). Therefore, it is useful to construe this influence as a direct arrow from the environment to AI systems *(embedding them in processes)*.

Information needs to flow from AI system governance to the organization in a bottom-up manner to maintain an overview of the organization's AI system portfolio. In turn, the organization can strive to proactively influence stakeholders and the normative regulatory and self-regulatory environment *(communication and engagement)*. Each level contains several AI governance components, which we explain in the following sections.

### 3.1. Environmental layer: Requirements from the contextual environment

An organization using AI faces several kinds of AI governance requirements originating from its contextual environment—that is, forces and factors that the organization cannot directly influence (Ramírez & Selsky, 2016). For simplicity, we have categorized the requirements into three types roughly in the order of most to least binding: hard law (normative regulation), principles and guidelines (self-regulation), and stakeholder pressure.

Hard law—that is, binding regulation for AI systems and AI user organizations—is currently under development, notably in the EU, where the EU AI Act was proposed in April 2021 (European Commission, 2021; Renda, 2020). The act will most likely come into force after rounds of negotiations and lobbying, similar to the legislative process of the General Data Protection Regulation (GDPR) (Minkkinen, 2019).

However, even before the enforcement of the EU AI Act, numerous data rules and application-specific and general rules are already relevant to AI systems (Viljanen & Parviainen, 2022). In other words, AI regulation exists, but its impacts remain underexplored, and more rules are emerging. Thus, the regulatory landscape is uncertain and fast-moving. This means that organizations must map regulations that are relevant to their AI systems and use cases.

AI ethics scholarship and practice have produced a plethora of self-regulatory documents outlining AI ethics principles. To make sense of this landscape, researchers have already produced numerous overviews of AI ethics guidelines (Hagendorff, 2020; Jobin et al., 2019; Schiff et al., 2020; Thiebes et al., 2021). A scoping review by Jobin et al. indicates convergence around the principles of transparency, justice and fairness, non-maleficence, responsibility, and privacy (Jobin et al., 2019). Despite some signs of convergence, organizations face often ill-defined and potentially contradictory guidelines, hence the need for organizational AI governance to operationalize into practice (Georgieva et al., 2022; Morley et al., 2020).

Stakeholder pressure on governing AI development and use is an emerging and currently understudied phenomenon. Citizen and consumer awareness of privacy issues has been researched (Steijn & Vedder, 2015; Tsohou & Kosta, 2017), but we were unable to find studies on the awareness of the novel domain of AI governance. Among the key stakeholders, investors generally pay increasing attention to environmental, social, and governance criteria when screening investments, but their awareness of AI governance issues remains limited (Minkkinen, Niukkanen, et al., 2022).

## 3.2. Organizational layer: Organizational practices and capabilities

The organizational layer of AI governance is composed of two key themes: strategic alignment and value alignment. These themes stem from the nature of AI governance as AI governance ensures the alignment of an organization's use of AI technologies with its organizational strategies and principles of ethical AI (Mäntymäki et al., 2022).

The first theme at the organizational level is *strategic alignment*, which requires the definition of an organizational strategy for the use of AI and then the alignment of this strategy with key organizational strategies (Eitel-Porter, 2021; Jöhnk et al., 2020). The organizational AI strategy provides a general direction and manages expectations regarding the overall set of AI systems that the organization intends to use and what they are meant to achieve. The strategically aligned use of AI requires organizational resources, capabilities, and processes that can be ensured, for example, through management commitment and staff training (Shneiderman, 2020).

Second, *value alignment* requires the organization's management to state the value base and AI ethics to which the organization adheres, and adherence to these values and principles across the organization's AI systems must be ensured. It also requires the adoption of a standpoint on tolerable regulatory, reputational, and other risks. Articulating the organization's risk tolerance is important because ethical deliberations are rarely clear-cut, and organizations need to manage trade-offs and tensions, for example, between efficiency and respect for privacy (Whittlestone et al., 2019).

A further aspect of value alignment is the consideration of the organization's AI systems' impacts on stakeholder groups, as in algorithmic impact assessment (Kaminski & Malgieri, 2020; Schiff et al., 2021). The inclusion of strategic and value alignments as separate themes highlights alignments at different levels. Most significantly, organizations need to align the use of AI with organizational strategies and objectives (Schneider et al., 2020) and with norms and ethical principles that emanate from the broader environmental layer (cf. Floridi, 2018)

## 3.3. AI system layer: Operational governance for the development, use, and management of AI systems

The AI system layer (operational governance for the development, use, and management of AI systems) refers to the operational level of AI governance. At this level, external requirements from the environment (environmental layer) and internal guidelines from management (organizational layer) are practically implemented in AI systems. Therefore, the organizational actors on the AI system layer differ from those on the organizational layer.

Instead of the management-level actors that are predominant on the organizational layer, the designers and developers of the AI systems play a key role in operational governance. The AI system layer is complex owing to the task of practically implementing functioning AI governance and the continuous advancement of AI technologies and governance tools. Thus, we present here an overview rather than an exhaustive account.

In the hourglass model of organizational AI governance, the AI system layer has eight components:

A. AI system
B. Algorithms
C. Data operations
D. Risk and impacts
E. Transparency, explainability, and contestability

F. Accountability and ownership
G. Development and operations
H. Compliance

The operational AI governance elements draw on several literature streams. They have been iteratively codeveloped and tested with AI practitioners and cross-checked against the EU AI Act proposal (European Commission, 2021) available at the time of writing.

*Table 1. Operational AI governance components, tasks, and literature streams*

| Governance components and tasks | Description | Literature streams |
|---|---|---|
| **A. AI system**<br>T1. AI system repository and AI ID<br>T2. AI system pre-design<br>T3. AI system use case<br>T4. AI system user<br>T5. AI system operating environment<br>T6. AI system architecture<br>T7. AI system deployment metrics<br>T8. AI system operational metrics<br>T9. AI system version control design<br>T10. AI system performance monitoring design<br>T11. AI system health check design<br>T12. AI system verification and validation<br>T13. AI system approval<br>T14. AI system version control<br>T15. AI system performance monitoring<br>T16. AI system health checks | Ensuring that the AI system is developed, operated, and monitored in alignment with the organization's strategic goals and values. | Software development and project management (Dennehy & Conboy, 2018) |
| **B. Algorithms**<br>T17. Algorithm ID<br>T18. Algorithm pre-design<br>T19. Algorithm use case design<br>T20. Algorithm technical environment design<br>T21. Algorithm deployment metrics design<br>T22. Algorithm operational metrics design<br>T23. Algorithm version control design<br>T24. Algorithm performance monitoring design<br>T25. Algorithm health check design<br>T26. Algorithm verification and validation<br>T27. Algorithm approval<br>T28. Algorithm version control<br>T29. Algorithm performance monitoring<br>T30. Algorithm health checks | Ensuring that the algorithms used by an AI system are developed, operated, and monitored in alignment with the organization's strategic goals and values. | Software development and project management (Dennehy & Conboy, 2018)<br>Critical algorithm studies (Kitchin, 2017; Ziewitz, 2016) |
| **C. Data operations**<br>T31. Data sourcing<br>T32. Data ontologies, inferences, and proxies<br>T33. Data pre-processing<br>T34. Data quality assurance<br>T35. Data quality metrics<br>T36. Data quality monitoring design<br>T37. Data health check design<br>T38. Data quality monitoring<br>T39. Data health checks | Ensuring that data are sourced, used, and monitored in alignment with the organization's strategic goals and values. | Data governance and data management (Abraham et al., 2019; Brous et al., 2016; Janssen et al., 2020)<br>Critical data studies (Iliadis & Russo, 2016) |
| **D. Risk and impacts**<br>T40. AI system harms and impacts pre-assessment<br>T41. Algorithm risk assessment<br>T42. AI system health, safety, and fundamental rights impact assessment<br>T43. AI system non-discrimination assurance<br>T44. AI system impact minimization<br>T45. AI system impact metrics design<br>T46. AI system impact monitoring design<br>T47. AI system impact monitoring<br>T48. AI system impact health check | Identifying, managing, and monitoring potential risks and impacts caused by the AI system to align the system with the organization's strategic goals and values. | Algorithmic impact assessment (Kaminski & Malgieri, 2020; Metcalf et al., 2021) |
| **E. Transparency, explainability, and contestability (TEC)**<br>T49. TEC pre-design<br>T50. TEC design<br>T51. TEC monitoring design<br>T52. TEC monitoring<br>T53. TEC health checks | Ensuring that the AI system transparency, explainability, and contestability is aligned with the organization's strategic goals and values. | Explainable AI (Barredo Arrieta et al., 2020; Laato, Tiainen, et al., 2022; Meske et al., 2022)<br>Algorithmic transparency (Ananny & Crawford, 2018; Wachter et al., 2017)<br>Contestability (Almada, 2019; Floridi, 2018) |
| **F. Accountability and ownership**<br>T54. Head of AI<br>T55. AI system owner<br>T56. Algorithm owner | Ensuring necessary decision rights and responsibilities to govern the AI system and its algorithmic components to align the system with the organization's strategic goals and values. | Algorithmic accountability (Martin, 2019; Shah, 2018; Wieringa, 2020)<br>IT governance (Brown & Grant, 2005; Gregory et al., 2018; Tiwana & Kim, 2015; Weill, 2008) |
| **G. Development and operations**<br>T57. AI development<br>T58. AI operations<br>T59. AI governance integration | Designing and implementing appropriate workflows and organizational structures for developing AI systems | Software development and project management (Dennehy & Conboy, 2018)<br>DevOps (Gall & Pigni, 2021)<br>MLOps (Mäkinen et al., 2021) |
| **H. Compliance**<br>T60. Regulatory canvassing<br>T61. Regulatory risks, constraints, and design parameter analysis<br>T62. Regulatory design review<br>T63. Compliance monitoring design<br>T64. Compliance health check design<br>T65. Compliance assessment<br>T66. Compliance monitoring<br>T67. Compliance health checks | Understanding the regulatory environment of an AI system and ensuring its compliance with the relevant regulations | Regulatory and legal studies (Kaminski & Malgieri, 2020; Viljanen & Parviainen, 2022) |

The processes under each operational AI governance component are described in more detail in the Appendix.

The governance components are framed as articulating and translating the organizational strategic goals and values in organizational processes. Consequently. AI governance is adaptive and follows the evolution of the requirements from the environmental layer. In summary, the different governance processes include knowledge production functions (e.g., system predevelopment and regulatory mapping), organizational design (e.g., decision structure design), and practical management and monitoring (e.g., risk and impact management and compliance monitoring).

The operational AI governance processes under each component in Table 1 are based on software development life cycle (SDLC) models (Laato, Mäntymäki, Islam, et al., 2022). The life cycle perspective is essential because AI systems are IT systems that need to be governed during their entire life cycles, from conception to operation and monitoring (Ibáñez & Olmeda, 2021; Laato, Birkstedt, et al., 2022; Laato, Mäntymäki, Minkkinen, et al., 2022; Raji et al., 2020).

*Fig. 3* presents a general life cycle model of AI system development formulated by the Organisation for Economic Co-operation and Development (OECD) AI expert group (OECD, 2019).

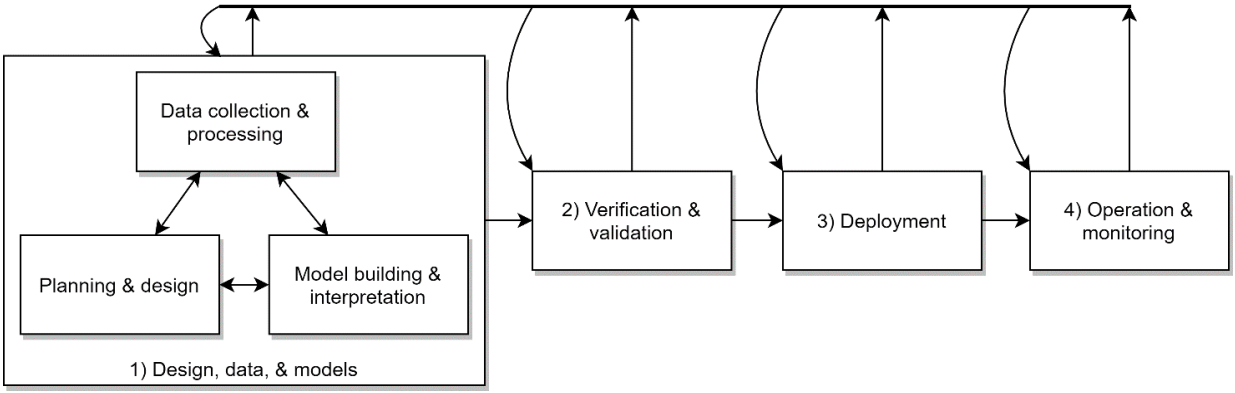

*Fig. 3. AI system life cycle (OECD, 2019)*

The eight operational AI governance components include processes at different stages of the AI system life cycle. *Fig.*4 presents the AI governance processes mapped to the four life cycle stages in the OECD model.

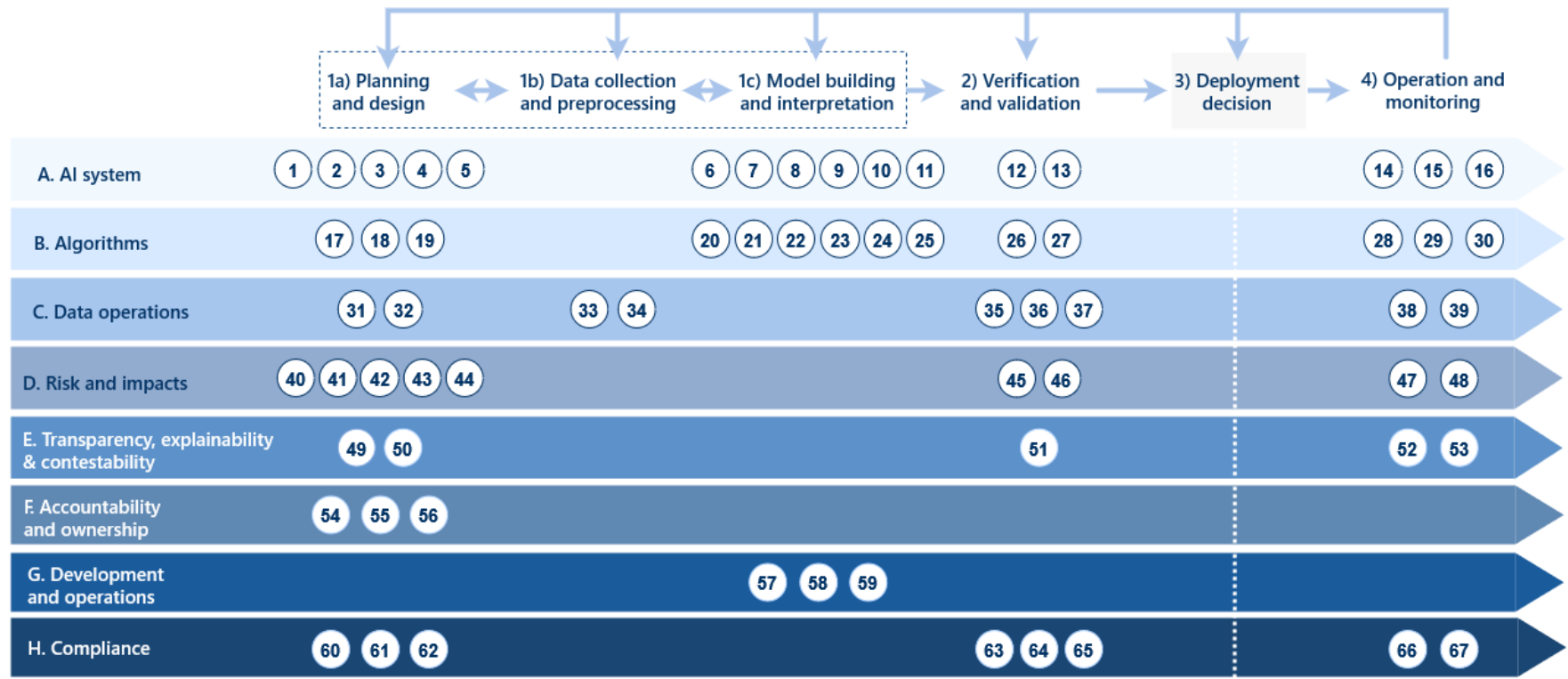

*Fig. 4. AI system life cycle and operational AI governance components*

Laying out the governance processes in the system life cycle shows the critical role of the design stage, where governance mainly affects knowledge production and organizational design. Toward the end of the AI system life cycle, requirements are associated with continuous management and monitoring when deploying and operating AI systems.

Six of the AI governance components (AI system, algorithms, data, risks and impact, TEC, and compliance) are structurally similar because they include processes in the design and model development, verification, deployment, and monitoring stages. From an AI governance perspective, accountability and development operations are more front-heavy governance areas. They require organizational efforts at the design stage, but they are not continuously managed and monitored in the same way as the previously mentioned six areas. It should be noted that this applies to accountability and development operations as parts of AI governance specifically and does not mean that these domains are absent from the later life cycle stages.

## 4. Discussion and Conclusion

### 4.1. Research implications

The chief research contribution of this paper is the integration of AI governance layers, components, and literature into an integrated model of organizational AI governance. These disparate elements have previously been discussed in various literature areas, as shown in the previous section.

Our integrated conception of AI governance is illustrated by the hourglass model (Fig. 2), which comprises three layers: 1) environmental requirements, 2) organizational practices and capabilities, and 3) operational governance of AI systems. Each layer points to different research fields. The environmental layer touches on technology and innovation policy (Stahl, 2022), AI ethics (Ibáñez & Olmeda, 2021), and regulatory studies (Kaminski & Malgieri, 2020; Viljanen & Parviainen, 2022), for instance. The organizational layer involves corporate governance (Harjoto & Jo, 2011), IT governance (Tiwana & Kim, 2015; Weill, 2008), and strategic management (Brendel et al., 2021), among others. Finally, the AI system layer involves fields such as software engineering (Dennehy & Conboy, 2018), critical algorithm studies (Kitchin, 2017; Ziewitz, 2016), and data governance (Abraham et al., 2019; Brous et al., 2016; Janssen et al., 2020). Together, the layers form a multidisciplinary area of study that researchers need to understand to enable them to unpack the challenges and phenomena surrounding AI governance.

Therefore, our research highlights the systemic, multilayered, and multi-stakeholder nature of AI governance (Minkkinen, Zimmer, et al., 2022; Zimmer et al., 2022). All aspects of AI governance, such as organizational structures, AI regulation, and AI system development practices, should be seen as interlinked components of a complex system. Going further, our research elucidates the need for a multidisciplinary understanding of AI governance, encompassing organizational studies, information systems, technology foresight, computer science, legal studies, and other fields. Examining these intertwined AI governance layers is likely to emerge as a significant topic in information systems, computer science, management and organization, and political science research (Mäntymäki et al., 2022).

### 4.2. Practical implications

The hourglass AI governance model is intended to help organizations implement AI governance and, in doing so, address the challenge of translating ethical AI principles into practice (Morley et al., 2020; Seppälä et al., 2021). However, this paper presents only an overview of the necessary AI governance processes, and further operationalization work is required from researchers and organizations. As part of the hourglass model, we have developed a list of operational AI governance processes (see Table 1). This description of each task is available in the Appendix.

AI governance is an emerging area of organizational practices, regulatory and ethical requirements, and commercial offerings. The hourglass model of organizational AI governance provides a starting point for organizational decision-makers (e.g., heads of AI and development team leads) to consider the governance components that need to be in place to ensure social acceptability, mitigate risks, and realize the potential of AI system deployment. In particular, the connection to AI system life cycles helps focus attention on pertinent issues at the right time in the AI system life cycle.

### 4.3. Future directions

Considering the specific ways forward for understanding systemic AI governance, we point to three particularly promising avenues for future research. First, in-depth and comparative case studies would elucidate how different organizations implement AI governance in practice. Empirical studies using sensitizing models, such as our hourglass model, would take AI governance scholarship beyond the currently predominant conceptual work. Second, the interplay and mediating mechanisms between the environmental, organizational, and operational AI governance layers could be fleshed out and conceptualized beyond our initial sketch. Third, the dynamics between AI governance actors, ranging from individuals and small companies to global players (e.g., the OECD), could be further investigated beyond the current global overviews (Butcher & Beridze, 2019; Johnson & Bowman, 2021; Schmitt, 2021). This research direction links to the recent literature that suggests that AI governance actors are organizing as ecosystems comprising companies, professional bodies, governmental organizations, and transnational entities such as the EU and OECD (Minkkinen, Zimmer, et al., 2022; Stahl, 2021).

**Acknowledgments**

The authors wish to thank Meeri Haataja and the Artificial Intelligence Governance and Auditing (AIGA) project consortium members for their valuable inputs during the development of the hourglass model. The authors also wish to thank Business Finland for funding this research.

# Appendix: AIGA AI Governance Framework

**Version February 2023**

The AIGA AI governance framework has been developed within the AIGA (AI governance and Auditing) research project and financially supported by the AI Business Program of Business Finland.

## Key premises of the AIGA AI Governance Framework

**The AIGA AI Governance Framework is a practice-oriented framework for implementing responsible AI.** With it, organizations can adopt a controlled, human-centric approach that covers the entire process of AI system development and operations.

**The AIGA AI Governance Framework supports compliance with the upcoming European AI regulation (the AI Act, under preparation).** This makes it relevant for organizations that perform in-house AI system development, particularly in high-risk application areas. Furthermore, the framework serves as a practical guide for any organization taking steps towards more responsible AI.

**The AIGA AI Governance Framework provides a template for decision-makers to address the key questions on the use of AI.** Organizations may use it to design and implement practices for using AI in a socially and ethically responsible manner.

**The AIGA AI Governance Framework is value-agnostic.** It does not give priority to any particular ethical stance but is designed to facilitate the development and deployment of transparent, accountable, fair, and non-maleficent AI systems.

# Table of contents

# A   AI system

## T1   AI system repository and AI ID

Coordinated AI development, operation, and use are essential to organizations' sustainable AI operations.

All organizations using AI systems should operate an AI system repository.

The repository should

1) identify all AI systems the organization is developing, operates, uses, or has retired,
2) assign them a unique identifier,
3) contain the relevant documents the organization has produced or received on the AI system.

## T2   AI system pre-design

Once an organization initiates an AI system development project, it should perform a preliminary pre-design of the system. The Head of AI (T54) should ensure that the organization

1) enters the AI system into the AI repository (T1),
2) assesses whether the AI system can align with the organization's values and risk tolerance,
3) initiates the development processes and assigns roles and responsibilities,
4) identifies and documents the planned AI system's key features and design constraints.

## T3   AI system use case

Identifying and understanding the intended use case of an AI system and its other possible uses is key to sustainable AI development and use. The use case affects the system's regulatory environment and may have significant reputational risk implications.

The AI System Owner (T55) should ensure that the organization defines and documents

1) the intended use case of the AI system and
2) the possible other uses of the AI system.

The AI System Owner should ensure that the use case definition aligns with the organization's values and risk tolerance.

The AI System Owner should ensure that the organization takes adequate measures to prevent inappropriate AI system misuse.

## T4   AI system user

People in organizations use AI systems. Some AI systems make decisions that directly or indirectly affect humans and their rights and obligations (affected persons). Sustainable AI system development and use require that the organization is conscious of who is using the AI system and whose rights and obligations it may affect.

The organization should define and document

1) the intended AI system user organizations and human users,
2) the intended affected persons, and
3) possible other users and affected persons.

The AI System Owner (T55) should ensure that the user definitions align with the organization's values and risk tolerance.

**T5**     **AI system operating environment**
AI systems are embedded in the business and organizational environment. This environment typically consists of technological and social elements. The operating environment is a key driver of AI system impacts.

The organization should define and document

1) the intended business or operational model and environment of the AI system,
2) the intended IT environment the AI system is embedded in and interacts with,
3) the other intended AI systems the AI system interacts with.

**T6**     **AI system architecture**
AI systems are part of IT systems. IT systems contain various data, computing infrastructures and resources, system architecture, and interfaces. The IT system architecture affects the AI system operations and may affect its risks and impacts.

The organization should

1) define and document the position of the AI system in the organization's IT architecture, and
2) document and manage the AI system's interactions with the organization's other IT systems.

**T7**     **AI system deployment metrics**
The organization can only ensure desired AI system performance by deploying appropriate metrics to evaluate it.

The AI System Owner should ensure that the organization defines and documents pre-deployment performance metrics that the AI system must meet before deployment or updates.

The AI System Owner should ensure that the performance metrics align with the organization's values and risk tolerance.

**T8**     **AI system operational metrics**
Organizations can only assure desired AI system performance by deploying appropriate metrics to evaluate it.

The AI System Owner should ensure that the organization defines and documents operational use performance metrics for assessing AI system performance during its operational use.

The AI System Owner should ensure that the key target performance metrics align with the organization's values and risk tolerance.

**T9**     **AI system version control**
AI systems will likely undergo several redesigns and update cycles during their lifetime. Designing and implementing an effective version control system integrated with the AI governance framework processes is crucial.

The AI System Owner should ensure that the organization defines, documents, and entrenches

1) quality control processes for new versions and updates and
1) version control and approval workflows.

The AI System Owner should ensure that the AI system version control design aligns with the organization's values and risk tolerance.

**T10  AI system performance monitoring design**

Monitoring AI system performance is crucial to ensuring that the system sustains the desired level of performance. The monitoring must be systematic and metrics-based to achieve consistency over time.

The AI System Owner should ensure that the organization defines, documents, and entrenches

1) workflows and technical interfaces to facilitate the monitoring of AI system performance, including, for example
   a) the automated or manual production and reporting of performance metrics data,
   b) alarm thresholds, and
   c) workflows that allocate monitoring responsibilities.
2) workflows and technical interfaces to facilitate the detection of system malfunctions and other anomalous events, and
3) workflows to address issues detected during health checks.

The AI System Owner should ensure that the AI system performance monitoring process aligns with the organization's values and risk tolerance.

**T11  AI system health check**

AI systems may be subject to performance deterioration over the medium and long term. In addition, the business, operational, IT, and regulatory environments and stakeholder pressures will change over time. These processes may jeopardize system performance or lead to unacceptable risks.

The AI System Owner should ensure that the organization conducts regular comprehensive reviews of the AI system (health checks) to ensure that the AI system aligns with the organization's values and risk tolerance.

The AI System Owner should ensure that the organization defines, documents, and entrenches workflows and technical interfaces for reviewing

1) AI system use case,
2) AI system users,
3) AI system operational environment,
4) AI system technical environment,
5) AI system metrics,
6) AI system version control practices,
7) the AI system monitoring practices and
8) the AI system health check practices.

**T12  AI system verification and validation**

Verifying and validating AI system performance is a crucial aspect of AI system development and quality control.

In AI systems with machine learning components, verification will require comprehensive validation testing and theoretical and analytical verification. In many cases, validation will require that the developer organization builds a simulation environment where it can explore algorithm performance using comprehensive samples of real-world, non-training data inputs.

The AI System Owner should ensure that the organization develops appropriate verification and validation methods for adequate AI system performance.

**T13  AI system approval**

Decisions on approving AI systems and AI versions of AI systems for operational use should be informed and preceded by a careful review of the AI system documentation.

Before deciding to deploy an AI system or AI system version, the AI System Owner should review all documentation on the AI system and ensure that the AI system impacts are acceptable and the system meets the performance targets for deployment.

### T14    AI system version control

The AI System Owner should ensure that the organization implements the planned AI system version control processes.

If the system version control processes disclose a breach of version control practices or indicate a value or risk tolerance, the AI System Owner should initiate appropriate measures to address the breach or regain alignment.

### T15    AI system performance monitoring

The AI System Owner should ensure that the organization implements the planned AI system performance monitoring processes.

If the system version control processes disclose a breach of performance standards or indicate a value or risk tolerance misalignment, the AI System Owner should initiate appropriate measures to address the breach or regain alignment.

### T16    AI system health checks

The AI System Owner should ensure that the organization performs the regular planned health checks.

The reviews should assess whether the AI system aligns with the organization's values and risk tolerance. If a review discloses a misalignment, the AI System Owner should initiate appropriate measures to regain alignment.

## B   Algorithms

### T17    Algorithm ID

Coordinated algorithm development, operation, and use are key to sustainable AI operations in organizations. While some algorithms are developed in-house and others procured from vendors, it is important that the organization is aware of the algorithms it is developing, operating, or using.

All organizations using AI systems should operate an Algorithm Repository.

The repository should

1) identify, to the extent possible, all algorithms the organization is developing, operates, uses, or has retired,
2) assign them a unique identifier,
3) contain the relevant documents the organization has produced or received on the algorithm.

### T18    Algorithm pre-design

Once an organization initiates an algorithm development project, it should perform a preliminary pre-design of the algorithm. The Head of AI should ensure that the organization

4) enters the algorithm into the Algorithm Repository,
5) assesses whether the algorithm can align with the organization's values and risk tolerance,
6) initiates the development processes and assigns roles and responsibilities,

7) identifies and documents the key features and design constraints for the planned algorithm.

**T19**  **Algorithm use case design**

Understanding the intended uses of an algorithm together with its possible misuses is key to sustainable AI development and use.

For each algorithm in its Algorithm Repository, the organization should define and document, to the extent possible,

1) the intended uses of the algorithm and
2) the possible foreseeable misuses of the algorithm, if relevant.

The use case definition should guide the development processes and build on the risk and impact pre-design and assessment outcomes.

The AI System Owner should ensure that the intended use case aligns with the organization's values and risk tolerance. The AI System Owner should ensure that the organization takes adequate measures to prevent inappropriate algorithm misuse.

**T20**  **Algorithm technical environment design**

When operational, algorithms are typically part of AI systems. The AI system architecture and its connections to the organization's other IT systems affect the AI system's impacts.

The organization should

1) define and document the position of the algorithm in the AI systems it is a part of,
2) document and manage interactions with the organization's other AI systems and IT systems.

The AI System Owner should ensure that the AI system's technical environment aligns with the organization's values and risk tolerance.

T21  **Algorithm deployment metrics design**

The organization can only ensure desired algorithm performance by designing appropriate metrics to evaluate it.

The Algorithm Owner (T56) should ensure that the organization defines and documents pre-deployment performance metrics that the algorithm must meet prior to deployment or updates.

The Algorithm Owner should ensure that the performance metrics align with the organization's values and risk tolerance.

T22  **Algorithm operational metrics design**

The organization can only ensure desired algorithm performance by designing appropriate metrics to evaluate it.

The Algorithm Owner should ensure that the organization defines and documents operational performance metrics for assessing algorithm performance during operational use.

The Algorithm Owner should ensure that the performance metrics align with the organization's values and risk tolerance.

T23  **AI system version control design**

AI system algorithms will likely undergo several redesigns and update cycles during their lifetime. Some algorithms may learn continually. Designing and implementing an effective version control system integrated with the AI governance framework processes is crucial to sustainable AI operations.

The AI System Owner should ensure that the organization defines, documents, and implements

3) quality control processes for new versions and updates and
4) version control and approval workflows.

The Algorithm Owner should ensure that the AI system version control design and practices align with the organization's values and risk tolerance.

## T24 Algorithm performance monitoring design

Monitoring algorithm performance is crucial to ensure that the organization sustains the desired level of operational performance. The monitoring must be systematic and metrics-based to achieve consistency over time.

The AI System Owner should ensure that the organization defines, documents, and implements

1) workflows and technical interfaces to facilitate the monitoring of AI system performance, including for example
2) automated or manual production and reporting of performance metrics data,
3) alarm thresholds, and
4) workflows that allocate monitoring responsibilities.
1) workflows to address issues detected during regular monitoring and health checks.

The Algorithm Owner should ensure that the AI system performance monitoring design process aligns with the organization's values and risk tolerance.

## T25 Algorithm health checks design

Algorithms may be subject to performance deterioration over the medium and long term. In addition, the business, operational, IT, and regulatory environments and stakeholder pressures will change over time. These processes may jeopardize algorithm performance or lead to the emergence of unacceptable risks.

The Algorithm Owner should ensure that the organization conducts regular comprehensive reviews of the algorithm (algorithm health checks) to ensure that the algorithm aligns with the organization's values and risk tolerance.

The Algorithm Owner should ensure that the organization defines, documents, and implements workflows and technical interfaces to review

1) the AI system use case,
2) the AI system users,
3) the AI system operational environment,
4) the AI system technical environment,
5) the AI system deployment metrics,
6) the AI system operational use metrics,
7) the AI system version control practices,
8) the AI system performance monitoring practices and
9) the AI system health check practices.

The reviews should assess whether the algorithm aligns with the organization's values and risk tolerance. If the review discloses misalignments, the Algorithm Owner initiates appropriate measures to regain alignment.

### T26  Algorithm verification and validation

Verifying and validating algorithm performance is a crucial aspect of AI system development and quality control.

In AI systems with machine learning components, verification will require comprehensive validation testing in addition to theoretical and analytical verification. In many cases, validation will require that the developer organization builds a simulation environment where it can explore algorithm performance using comprehensive samples of real-world, non-training data inputs. Further, validation may require developing post hoc interpretability tools to gain insight into algorithm logic.

The Algorithm Owner should ensure that the organization develops appropriate verification and validation methods to ensure adequate algorithm performance.

### T27  Algorithm approval

Decisions approving algorithms and algorithm versions for operational use should be informed and preceded by a careful review of the algorithm.

Prior to deciding to deploy an algorithm or algorithm version, the AI owner should review all documentation on the algorithm and associated and ensure that the algorithm meets the performance targets for deployment.

At times, the organization may have limited access to the algorithms in its AI system. In these cases, the approval process should review all available documentation and make a decision on whether deploying the algorithm creates risks that do exceed the organization's risk tolerance or breach its legal obligations.

### T28  Algorithm version control

The Algorithm Owner should ensure that the organization implements the planned AI system version control processes.

If the system version control processes disclose a breach of version control practices or indicate a value or risk tolerance, the AI System Owner should initiate appropriate measures to address the breach or regain alignment.

### T29  Algorithm performance monitoring

The Algorithm Owner should ensure that the organization implements the planned algorithm performance monitoring processes.

If the performance monitoring processes disclose a breach of performance standards or indicate a value or risk tolerance misalignment, the AI System Owner should initiate appropriate measures to address the breach or regain alignment.

### T30  Algorithm health checks

The Algorithm Owner should ensure that the organization performs the regular planned health checks.

The reviews should assess whether the AI system aligns with the organization's values and risk tolerance. If a review discloses a misalignment, the AI System Owner should initiate appropriate measures to regain alignment.

# C Data operations

### T31 Data sourcing

Data is crucial to both AI systems and algorithm development and operations.

The AI System Owner Sources should ensure that the organization defines and documents AI system data sources. The AI System Owner should ensure that the organization has the right to process the data.

The Algorithm Owner should ensure that the organization defines and documents training, validation, and operational data sources. The Algorithm Owner should ensure that the organization has the right to process the algorithm training, validation, and operational data.

### T32 Data ontologies, inferences, and proxies

Data resources contain various categories of data. The categories reflect explicit or implicit data ontologies. Data ontologies consist of entity taxonomies (what entities are assumed to exist) and models of entity relationalities and causality (how the entities relate to each other). Data ontologies may have significant implications on how algorithms and AI systems function, what risks they create and to whom, and what entities and how the AI system affects them. In advanced machine learning approaches, data ontologies are complex as the source data ontologies combine with the non-representational sensemaking inherent to the approaches. Understanding the ontologies may only be possible by analyzing algorithm outputs.

The AI System Owner should ensure that the organization

10) adequately understands the AI system data ontology,
11) has explored the risks related to possible inconclusive evidence, system, and discrimination risk, and
12) develops and implements measures to minimize and mitigate possible data-related risks.

The Algorithm Owner should ensure that the organization

1) adequately understands the algorithm data ontology,
2) adequately understands what inferences are drawn on the data and what proxies are created when the organization uses a machine learning approach to develop an algorithm,
3) has explored the risks related to possible inconclusive evidence, system bias, and discrimination risks the data ontology may create, and
4) develops and implements measures to minimize and mitigate possible data ontology-related risks.

In particular, if the AI system makes decisions that affect natural persons, the AI system owner should ensure that the organization conducts a comprehensive assessment of the AI system's discrimination, misidentification, and cultural sensitivity risks.

The AI System Owner and Algorithm owner should ensure that the residual risks are acceptable and align with the organization's values and risk tolerance.

### T33 Data preprocessing

Training, testing, and operation undergo preprocessing in many AI systems.

The AI System Owner and Algorithm Owner should ensure that the organization designs and implements appropriate workflows and technical interfaces for effective and appropriate data preprocessing.

As training data, validation data, and operational data often differ qualitatively, the Algorithm Owner should ensure that the organization understands the differences and designs and implements appropriate workflows and interfaces for preprocessing each data category.

The AI System Owner should ensure that the data preprocessing process aligns with the organization's values and risk tolerance

### T34   Data quality assurance

Adequate data quality is a crucial precondition to all AI system operations.

The AI System Owner should ensure that the organization designs and entrenches appropriate workflows and technical arrangements for

1) gathering and producing information on data quality, and
2) ensuring that the data (including the training, validation, and testing data) is of adequate quality and sufficiently relevant, complete, and representative.

Data quality analyses should also include an analysis of additional data needs.

### T35   Data quality metrics

The organization can only ensure desired AI system performance by designing appropriate metrics to evaluate data quality.

The AI System Owner should ensure that the organization defines and documents data quality metrics for assessing the quality of the data the AI system uses.

The Algorithm Owner should ensure that the performance metrics align with the organization's values and risk tolerance.

### T36   Data quality monitoring design

Monitoring AI system data quality is crucial to ensuring that the AI system sustains the desired level of operational performance. Data quality monitoring must be systematic and metrics-based to achieve consistency over time.

The AI System Owner should ensure that the organization defines, documents, and entrenches workflows and technical interfaces to facilitate the monitoring of data quality. In particular, the AI System Owner should identify anomalous data entries and data drift. including

1) automated or manual production and reporting of data quality indicators, alarm thresholds, and
2) workflows that allocate monitoring responsibilities.
3) workflows to address issues detected during regular monitoring.

The Algorithm Owner should ensure that the data quality design process aligns with the organization's values and risk tolerance.

### T37   Data health check design

Data resources may be subject to deterioration over the medium and long term. In addition, the business, operational, IT, and regulatory environments and stakeholder pressures will change over time. These processes may jeopardize data access or data quality and lead to unacceptable risks.

The AI System Owner should ensure that the organization designs processes for regular comprehensive reviews of the AI system resources (Data health checks) to ensure that the data-related risks are acceptable and align with the organization's values and risk tolerance.

The AI System Owner should ensure that the organization defines, documents, and entrenches workflows and technical interfaces to review

1) AI system and algorithm data sources,
2) data preprocessing practices,
3) data quality, and
4) data ontology, inferences, and proxies.

**T38    Data quality monitoring**
The AI System Owner should ensure that the organization implements the planned data quality processes.

If the data quality control processes disclose a breach of data quality standards, data drift, or indicate a value or risk tolerance misalignment, the AI System Owner should initiate appropriate measures to address the breach or regain alignment.

**T39    Data health checks**
The Algorithm Owner should ensure that the organization performs the regular planned health checks.

The reviews should assess whether the AI system data resources and data-related processes align with the organization's values and risk tolerance. If a review discloses a misalignment, the AI System Owner should initiate appropriate measures to regain alignment.

# D  Risk and impacts

### T40  AI system harms and impacts pre-assessment

Understanding what harms and societal impacts an AI system may create is a crucial precondition for sustainable AI system development. The intensity of potential harms and impacts will vary significantly across different AI systems. Industrial AI systems with no direct effects on individuals or the environment will likely create limited harms and impacts. An AI system that makes irreversible decisions that affect the rights and obligations of individuals will have profound impacts and may generate significant harms.

The AI System Owner should ensure that the organization conducts a harms and impact pre-assessment at the outset of AI system development. The pre-assessment outcomes should be documented.

The pre-assessment should cover a wide range of potential harm and impact creation pathways. The pre-assessment should consider the harms the AI system may create and the impact it may have on its users, possible decision-making subjects, other affected parties, society at large, and the environment.

AI system risk assessments often focus on the direct harms the systems may create. The harms and impacts pre-assessment should, however, aim at also identifying the potential system-level harms and impacts. These include the social action affordances the system may create or modify, its potential wealth and power distribution implications and effects on equality.

The ethical advisory function should be involved in the pre-assessment of the harms and impacts if it is likely that the AI system will create a non-negligible risk of harm to individuals or the environment.

The AI System Owner should ensure that harms and impacts pre-assessment is conducted if the design parameters of the AI system undergo fundamental changes.

### T41  Algorithm risk assessment

Algorithms constitute the backbones of AI systems. AI system performance driven by algorithm performance. Possible AI systems biases and unfair outcomes often emanate from algorithm design. If the organization has access to the algorithms in the AI systems, identifying the possible algorithm risks and assessing their gravity is key to sustainable AI system development and operation.

Algorithm risk assessment should cover, to the extent possible, a wide range of algorithm-related risk sources and causes.

The Algorithm Owner should at least ensure that the organization

1) explores and documents how the algorithm affects the operations of the entire AI system
2) explores and documents the possible risk of biased and, in particular, discriminatory outcomes, and
3) explores and documents the risk of unfair outcomes and harms the algorithm may generate.

As identifying biases and unfairness is often complex and contentious, the reviews should involve ethical and legal experts. Particularly if the organization intends to use the algorithm in a high-risk use case. In machine learning algorithms, testing algorithm outputs may be necessary for identifying biases and discriminatory outcomes.

In addition, if a machine learning algorithm incorporates inferences made from training data, the risk assessment should review and assess

4) the risk of detecting non-existing patterns and correlations in the data,
5) the level of algorithmic scrutability and explainability.

### T42 AI system health, safety, and fundamental rights impact assessment

AI system impacts on the health and safety of humans will likely remain the most important concerns that organizations should address when developing and using AI. These impacts together with fundamental rights impacts will likely be the centerpieces of future regulatory instruments.

If the AI system harms and impacts pre-assessment (T40) indicates that the AI systems will likely have non-negligible impacts on the health, safety, and fundamental rights of individuals, The AI System Owner should ensure that the organization undertakes and documents

6) a health impact review to identify the potential health impacts the AI system may have on the physical and psychological well-being of its users, subjects, and other affected parties,
7) a safety impact review to the identify the potential safety risks the AI system may impose on individuals' and organizations' tangible assets, and
8) a fundamental rights impact to identify the potential impacts that the AI system may have on the protection and realization of individuals' fundamental rights. The legal advisory function should be involved in the assessment.

### T43 AI system non-discrimination assurance

Many jurisdictions have non-discrimination laws and impose equal treatment requirements. Non-compliance with the non-discrimination laws and equal treatment requirements is incompatible with sustainable AI operations and may create significant legal and reputational risks.

The AI System Owner should ensure that the organization conducts and documents a non-discrimination assurance process to ensure that the AI system outputs are compliant with non-discrimination laws and equal treatment requirements. The legal advisory function should be involved in both designing and conducting the assurance.

Ensuring that an AI system creates no discrimination risk is challenging due to the nature of the non-discrimination and equal treatment. For example, under the Finnish Equality Act, an AI system would directly discriminate against a person if the system threated a person less favorably than other based on their age, nationality, language, religion, belief, opinion, political activity, trade union activity, family relationships, state of health, disability, sexual orientation, or other personal characteristics. Less favorable treatment is discrimination even if based on an apparently neutral rule. Despite the prima facie ban, differential treatment can be justified if mandated by law or the treatment has an acceptable objective in terms of basic and human rights, and the measures to attain the aim are proportionate.

Conducting a diligent non-discrimination assurance is particularly important for AI systems with algorithms developed using machine learning approaches. Machine learning approaches may result in inadvertent discrimination. As the algorithms are often unexplainable, detecting discriminatory bias may require the use of post-hoc analysis tools and real-world data AI system output testing.

### T44 AI system impact minimization

Minimizing the AI system impacts is an important phase in sustainable AI system development and deployment. Minimizing the impacts requires first that the potential impacts are analyzed and appropriate measures are taken to eliminate or reduce adverse impacts where possible. Second, minimization requires that the organization mitigates the effects of the adverse impacts that it cannot eliminate or, third, manages their consequences.

To arrive at an acceptable AI system impact, the AI System Owner should ensure that the organization

1) conducts a thorough analysis of potential impacts the system may have on its users, subjects or affected parties, or the environment,
2) develops and implements a risk minimization plan.

The risk minimization plan should be designed to guarantee that the AI systems are acceptable and aligned with the organization's values and risk tolerance.

The risk minimization plan should outline

1) appropriate measures to eliminate adverse impacts to the extent possible,
2) appropriate measures to reduce adverse impacts that cannot be eliminated,
3) appropriate measures to mitigate the effects of the residual adverse impacts, and
4) appropriate measures to manage the adverse impacts that cannot be mitigated.

### T45   AI system impact metrics design

Acceptable AI system impacts performance can only be ensured by deploying appropriate metrics to measure them.

The AI System Owner should ensure that the organization defines and documents metrics for monitoring the AI system impacts during its operational use.

The AI System Owner should ensure that the impact metrics align with the organization's values and risk tolerance.

### T46   AI system impact monitoring design

Monitoring AI impact is crucial to ensuring that its impacts remain acceptable. The monitoring must be systematic and metrics-based to achieve consistency over time.

The AI System Owner should ensure that the organization defines, documents, and entrenches

5) workflows and technical interfaces to facilitate the monitoring of AI system impact, including for example
6) automated or manual production and reporting of impact metrics data,
7) alarm thresholds, and
8) workflows that allocate monitoring responsibilities.
9) workflows to address issues detected during health checks.

The AI System Owner should ensure that the AI system performance monitoring process aligns with the organization's values and risk tolerance.

### T47   AI system impact monitoring

The AI System Owner should ensure that the organization implements the planned AI system impact monitoring processes.

If the system version control processes disclose a breach of impact standards or indicate a value or risk tolerance misalignment, the AI System Owner should initiate appropriate measures to address the breach or regain alignment.

### T48   AI system impact health check

The AI System Owner should ensure that the organization performs the regular planned impact health checks.

The reviews should assess whether the AI system impacts align with the organization's values and risk tolerance. If a review discloses a misalignment, the AI System Owner should initiate appropriate measures to regain alignment.

# E   Transparency, explainability, contestability

### T49   TEC expectation canvassing

AI systems will be subject to varying stakeholder transparency, explainability, and contestability requirements.

Some of the expectations may be regulatory in origin. GDPR, for example, imposes a right to explanation for automated decisions.

Internal and external stakeholders also impose TEC requirements on the system. For example, recommendation engine users may require information on the system logic to be able to assess the accuracy and appropriateness of the recommendations.

To prevent investment slippage, the AI System Owner should ensure that the organization conducts a TEC requirement canvassing at the outset of the development process. In the canvassing, the organization should identify relevant TEC stakeholders and TEC expectations.

Identifying regulatory transparency and explainability expectations typically requires input from legal experts and is a part of the compliance process. Identifying non-regulatory TEC stakeholder expectations may benefit from stakeholder consultations or co-creation efforts.

### T50   TEC design

After identifying the appropriate transparency, explainability, and contestability expectations, the AI System Owner should ensure that the organization designs appropriate technical and organizational structures to satisfy the TCE expectations and align the AI system with the organization's values and risk tolerance.

The TCE design should address how a sufficient level of transparency, explainability, and contestability is secured by

1) algorithm and AI system design,
2) technical interfaces and other arrangements that allow the end-users and others affected to gain adequate visibility into the AI system,
3) procedures, technical systems, technical interfaces, and other arrangements that allow decision-making subjects to contest the decisions in an appropriate manner and to an appropriate degree, and
4) user instructions and end-user transparency and explainability documents such as explanations of decision-making logic.

### T51   TEC monitoring design

Monitoring AI system TCE performance is crucial to ensuring that the organization sustains the desired level of AI system performance. The monitoring must be systematic and metrics-based to achieve consistency over time.

The AI System Owner should ensure that the organization defines, documents, and entrenches

1) workflows and technical interfaces to facilitate the monitoring of AI system TEC performance, including for example
2) automated or manual production and reporting of performance metrics data,
3) alarm thresholds, and
4) workflows that allocate monitoring responsibilities.
5) workflows to address issues detected during regular monitoring and health checks.

The AI System Owner should ensure that the AI system TEC performance monitoring process aligns with the organization's values and risk tolerance.

**T52 TEC monitoring**

The AI System Owner should ensure that the organization implements the planned TEC monitoring processes.

If the system version control processes disclose a breach of TEC standards or indicates a value or risk tolerance misalignment, the AI System Owner should initiate appropriate measures to address the breach or regain alignment.

**T53 TEC health checks**

The AI System Owner should ensure that the organization implements planned TEC health monitor measures.

If the system monitoring discloses a breach of TCE performance standards or indicates a value or risk tolerance misalignment, the AI System Owner should initiate appropriate measures to address the breach or regain alignment.

# F  Accountability and ownership

### T54  Head of AI

Coordinated and accountable AI development, operation, and use are key to sustainable AI operations in organizations.

The organization should establish an organizational role responsible for overseeing AI system development and AI system operations. We refer to this organizational role as Head of AI.

The Head of AI should have sufficient knowledge and understanding of the organization's AI operations to make informed risk-reward decisions.

The Head of AI should have adequate authority and resources to implement and entrench coordinated AI operations.

### T55  AI system owner

Accountable AI development, operation, and use are key to sustainable AI operations in organizations.

Each AI system should have an organizational owner who oversees AI system development, deployment, operations, and retirement decisions. We refer to this organizational role as AI System Owner.

The AI System Owner should have sufficient knowledge and understanding of the AI system to make informed risk-reward decisions.

The AI System Owner should have adequate authority and resources to ensure that the AI system aligns with the organization's values and risk tolerance.

### T56  Algorithm owner

Accountable algorithm development, operation, and use are key to sustainable AI operations in organizations. While some algorithms are developed in-house and others procured from vendors, it is important that the organization has a designated person accountable for the algorithm.

Each algorithm, to the extent possible, should have an organizational owner who oversees algorithm development, deployment, operations, and retirement decisions in cooperation with the AI System Owners and the Head of AI. We refer to this organizational role as Algorithm Owner.

The Algorithm Owner should have sufficient knowledge and understanding of the algorithm to make informed risk-reward decisions.

The Algorithm Owner should have adequate authority and resources to ensure that the AI system aligns with the organization's values and risk tolerance.

# G  Development and operations

### T57  AI development

Coordinated and accountable AI development operations is key to sustainable AI development in organizations.

The Head of AI should ensure that the organization designs and implements

6) appropriate workflows and processes for its AI-related data acquisition, permitting, and analytics operations, including approvals and signoffs, and
7) appropriate workflows and processes for its algorithm development and AI system development operations, including approvals and signoffs.
8) The workflows and processes should align with the organization's values and risk tolerance.

### T58  AI operations

Coordinated and accountable AI operations are key to sustainable AI use in organizations.

The Head of AI should ensure that the organization designs and implements

1) appropriate workflows and processes for AI system operations, including approvals and signoffs, and
2) appropriate workflows and processes for AI retirement, including approvals and signoffs.
3) The workflows and processes should align with the organization's values and risk tolerance.

### T59  AI governance integration

AI systems are tools that serve business or operational purposes. AI system governance overlaps with business and operational, data, IT system, and governance and strategic governance. AI system governance should be integrated and aligned with other organizational governance processes.

The Head of AI should ensure that AI governance is integrated with other organizational governance processes.

# H Compliance

### T60 Regulatory canvassing

AI systems are typically subject to a variety of regulatory instruments that may force particular design choices, constrain functionalities, or in extreme cases make implementing a specific design, use case, or business model impossible. Understanding the regulatory environment is, consequently, important to prevent misplaced investments.

To develop a preliminary understanding of the AI system regulatory environment, the AI System Owner should ensure that the organization conducts a regulatory environment canvassing. The regulatory canvassing provides the organization with basic information on the regulatory environment in which the AI system will be used.

The canvassing should ensure that laws and regulations that may affect the AI system are identified and reviewed. The canvassing should develop a knowledge base of the contents of the primary regulatory instruments and key constraints that could affect AI system design and operations. All parties active in developing or implementing an AI system within the organization should be aware of the findings of the canvassing process.

### T61 Regulatory risks, constraints, and design parameter analysis

Regulation may impose critical constraints and requirements on AI system design.

Once a tentative understanding of the future intended use case and users of the AI system is reached, the AI System Owner should ensure that the legal function conducts an in-depth analysis of the system and its regulatory environment to identify key regulatory risks, constraints, and design parameters.

The analysis should

4) assess regulatory risks associated with known design options,
5) identify key design constraints,
6) identify design areas with significant regulatory implications (key regulatory issues), and
7) outline possible design options and their implications.

These regulatory focal points should be clearly communicated to all parties active in developing or implementing an AI system within the organization.

### T62 Regulatory design review

Development investments may be lost if the AI system has non-compliant features. To ensure efficient resource allocation and prevent investment slippage, the legal function should be consulted prior to making design decisions that could affect the key regulatory focal points.

the AI System Owner should ensure that developers consult the legal function before significant decisions affecting key regulatory issues are made.

### T63 Compliance monitoring design

The regulatory environment will likely change during AI system lifetime. Changes to the AI system may also disrupt compliance.

The AI System Owner should ensure that the organization maintains an awareness of possible regulatory changes relevant to the AI system.'

The organization should develop and entrench appropriate workflows and technical interfaces to facilitate compliance monitoring.

**T64 Compliance health check design**

The regulatory environment will likely change during the AI system's lifecycle. Changes to the AI system may also disrupt compliance.

The AI System Owner should ensure that the organization conducts regular comprehensive reviews of AI system compliance.

The organization should develop and entrench appropriate workflows and technical interfaces to facilitate periodic compliance reviews.

**T65 Compliance assessment**

The AI System Owner must ensure that the legal function conducts and documents a compliance assessment before the AI system is approved for operational use or a materially new version is deployed.

**T66 Compliance monitoring**

The AI System Owner should ensure that the organization implements the planned AI system compliance monitoring processes.

If the system version control processes disclose a non-compliance event, the AI System Owner should initiate appropriate measures to address the breach or regain alignment.

**T67 Compliance health checks**

The AI System Owner should ensure that the organization performs the regular planned compliance health checks.

The reviews should assess whether the AI system compliance processes align with the organization's values and risk tolerance. If a review discloses a misalignment, the AI System Owner should initiate appropriate measures to regain alignment.